\documentclass[conference]{IEEEtran}
\IEEEoverridecommandlockouts
\usepackage{amsmath,amssymb,amsfonts}
\usepackage{algorithm}
\usepackage{algpseudocode}

\usepackage{graphicx}
\usepackage{textcomp}
\usepackage{multirow}
\usepackage{booktabs}
\usepackage[dvipsnames,svgnames,table, x11names]{xcolor}

\usepackage{hyperref}     % for clickable links
\usepackage{tikz}
\usepackage{amsmath}
\usepackage{amssymb}
\usetikzlibrary{shapes.geometric, arrows.meta, positioning, fit, backgrounds, calc}
\usepackage{booktabs}
\usepackage{multirow}
\usepackage{array}
\usepackage{tabularx}
\usepackage[absolute,overlay]{textpos} % for textblock

\usepackage{placeins} % optional
\usepackage{pifont}

\definecolor{inputcolor}{RGB}{74, 144, 226}
\definecolor{backbonecolor}{RGB}{103, 191, 92}
\definecolor{contrastcolor}{RGB}{237, 106, 90}
\definecolor{fusioncolor}{RGB}{255, 159, 64}
\definecolor{losscolor}{RGB}{153, 102, 255}
\definecolor{outputcolor}{RGB}{201, 203, 207}

\tikzset{
    inputbox/.style = {
        rectangle, 
        rounded corners, 
        minimum width=4cm,  % Fixed width
        minimum height=1cm, 
        text centered, 
        draw=black, 
        fill=inputcolor!30, 
        font=\small\bfseries
    },
    backbone/.style = {
        rectangle, 
        rounded corners, 
        minimum width=3cm,  % Same fixed width
        minimum height=1cm, 
        text centered, 
        draw=black, 
        fill=backbonecolor!30, 
        font=\small
    },
    module/.style = {
        rectangle, 
        rounded corners, 
        minimum width=4cm,  % Same fixed width
        minimum height=1cm, 
        text centered, 
        draw=black, 
        fill=contrastcolor!30, 
        font=\small
    },
    fusion/.style = {
        rectangle, 
        rounded corners, 
        minimum width=3cm,  % Same fixed width
        minimum height=1cm, 
        text centered, 
        draw=black, 
        fill=fusioncolor!30, 
        font=\small
    },
    loss/.style = {
        rectangle, 
        rounded corners, 
        minimum width=3cm,  % Same fixed width
        minimum height=1cm, 
        text centered, 
        draw=black, 
        fill=losscolor!30, 
        font=\footnotesize
    },
    output/.style = {
        rectangle, 
        rounded corners, 
        minimum width=4cm,  % Same fixed width
        minimum height=1cm, 
        text centered, 
        draw=black, 
        fill=outputcolor!30, 
        font=\small\bfseries
    },
    arrow/.style = {thick,->,>=Stealth},
    line/.style = {thick,-},
    darrow/.style = {thick,<->,>=Stealth, dashed},
    annotation/.style = {
        rectangle, 
        draw=none, 
        fill=none, 
        font=\tiny, 
        text width=2.5cm, 
        align=center
    }
}

\hypersetup{
    colorlinks=true,      % enable colored links
    linkcolor=green,        % color for internal links
    citecolor=blue,        % color for citations
    urlcolor=Green          % color for external links (like your GitHub URL)
}

\def\BibTeX{{\rm B\kern-.05em{\sc i\kern-.025em b}\kern-.08em
    T\kern-.1667em\lower.7ex\hbox{E}\kern-.125emX}}

\usepackage{fancyhdr}
\begin{document}

\title{\texttt{xModel-KD}: Cross-modal Knowledge Distillation for 3D Scene Perception using LiDAR
\thanks{We acknowledge that this research was enabled in part by the support of the Natural Sciences and Engineering Research Council of Canada (NSERC), DGECR/416-2022, and the support provided by the Digital Research Alliance of Canada (https://alliancecan.ca/en).  }
}

\author{
\IEEEauthorblockN{Thenukan Pathmanathan}
\IEEEauthorblockA{\textit{Dept. of Computer Science} \\
\textit{Lakehead University} \\
Thunder Bay, Canada \\
tpathma1@lakeheadu.ca}
\and
\IEEEauthorblockN{Kanchan Keisham}
\IEEEauthorblockA{\textit{School of Computer Science Engg. \& Info. Systems} \\
\textit{Vellore Institute of Technology}\\
Tamil Nadu, India\\
kanchankeisham@gmail.com}
\and
\IEEEauthorblockN{Akilan Thangarajah}
\IEEEauthorblockA{\textit{Dept. of Software Engg.} \\
\textit{Lakehead University} \\
Thunder Bay, Canada \\
takilan@lakeheadu.ca
}

}

\maketitle

\begin{textblock*}{21cm}(0cm,0.5cm)
% Apply the same style to the first page
\begin{center}
    {\small\textit{Accepted in The 23rd Conference on Robots and Vision (CRV 2026)}}
\end{center}

\end{textblock*}

\begin{abstract}
% Point cloud segmentation is a fundamental task for 3D scene understanding, but remains notoriously data-hungry due to the high cost and time required for dense 3D annotations. A key challenge lies in the inherent limitations of individual sensing modalities: while 2D images provide rich texture and appearance cues, they lack explicit distance and geometric information; conversely, 3D point clouds capture accurate spatial structure and depth but are sparse and devoid of texture information. Relying on either modality alone, therefore, limits representation quality and generalization. 
Point cloud segmentation is a fundamental task in 3D scene understanding. Its progress is constrained by the high cost and time required for dense 3D annotations, making labeled samples difficult to obtain. Beyond annotation scarcity, different sensing modalities face inherent limitations. 2D images provide rich texture and appearance cues, yet they lack explicit depth and geometric structure. In contrast, 3D point clouds capture accurate spatial geometry but are sparse and contain no texture information. As a result, relying on a single modality restricts the richness of learned representations and weakens generalization.
Although recent multi-modal methods that combine 3D point clouds with 2D images have demonstrated strong performance in tasks such as classification and retrieval, they typically depend on large-scale labeled datasets and have not been fully exploited for data-efficient dense prediction. 
To address these limitations, we propose a novel cross-modal knowledge distillation framework, \texttt{xModel-KD}, for 3D point cloud segmentation. Our method exploits the complementary strengths of 2D texture and 3D geometry by learning unified per-point representations through cross-modal alignment.
Specifically, we design a cross-modal fusion encoder trained with a contrastive objective that enforces feature consistency between corresponding 2D and 3D representations across multiple views. By integrating powerful pre-trained backbones with a targeted fusion strategy, the proposed framework effectively transfers appearance cues from images to geometry-aware point features.
Experimental results show that cross-modal fusion achieves a $2\%$ absolute improvement in mIoU over a LiDAR-only baseline, demonstrating the benefit of leveraging complementary multi-modal information for scalable and annotation-efficient 3D scene understanding. \href{https://github.com/thenukan/xModel-KD}{Source code is available here.}
\end{abstract}


\section{Introduction}
Semantic segmentation of outdoor and indoor scenes is a fundamental task for autonomous driving and robotics, enabling robust scene understanding and safe navigation~\cite{Behley2019SemanticKITTI, caesar2020nuscenes}.
Recent years have witnessed remarkable progress in both camera-based methods \cite{wu2019pointconv, lin2016piecewise, huang2019ccnet} and LiDAR-based methods \cite{tang2020spvconv, zhu2021cylinder3d}.
While cameras provide rich appearance cues (color, texture, and fine boundaries), they lack explicit depth information and are sensitive to illumination changes.
In contrast, LiDAR point clouds provide accurate geometric structure and robustness to environmental conditions, but lack semantic richness and exhibit non-uniform point density with increasing distance.
These complementary properties have motivated emerging research on multi-modal solutions \cite{zhuang2021perception, li2023mseg3d, yan2022_2dpass}, where image features are injected into point cloud information via point-to-pixel correspondences.

\begin{figure}[!tp]
    \centering
    \includegraphics[trim={0.5cm, 3cm, 0.5cm, 0.5cm}, clip, width=1\columnwidth]{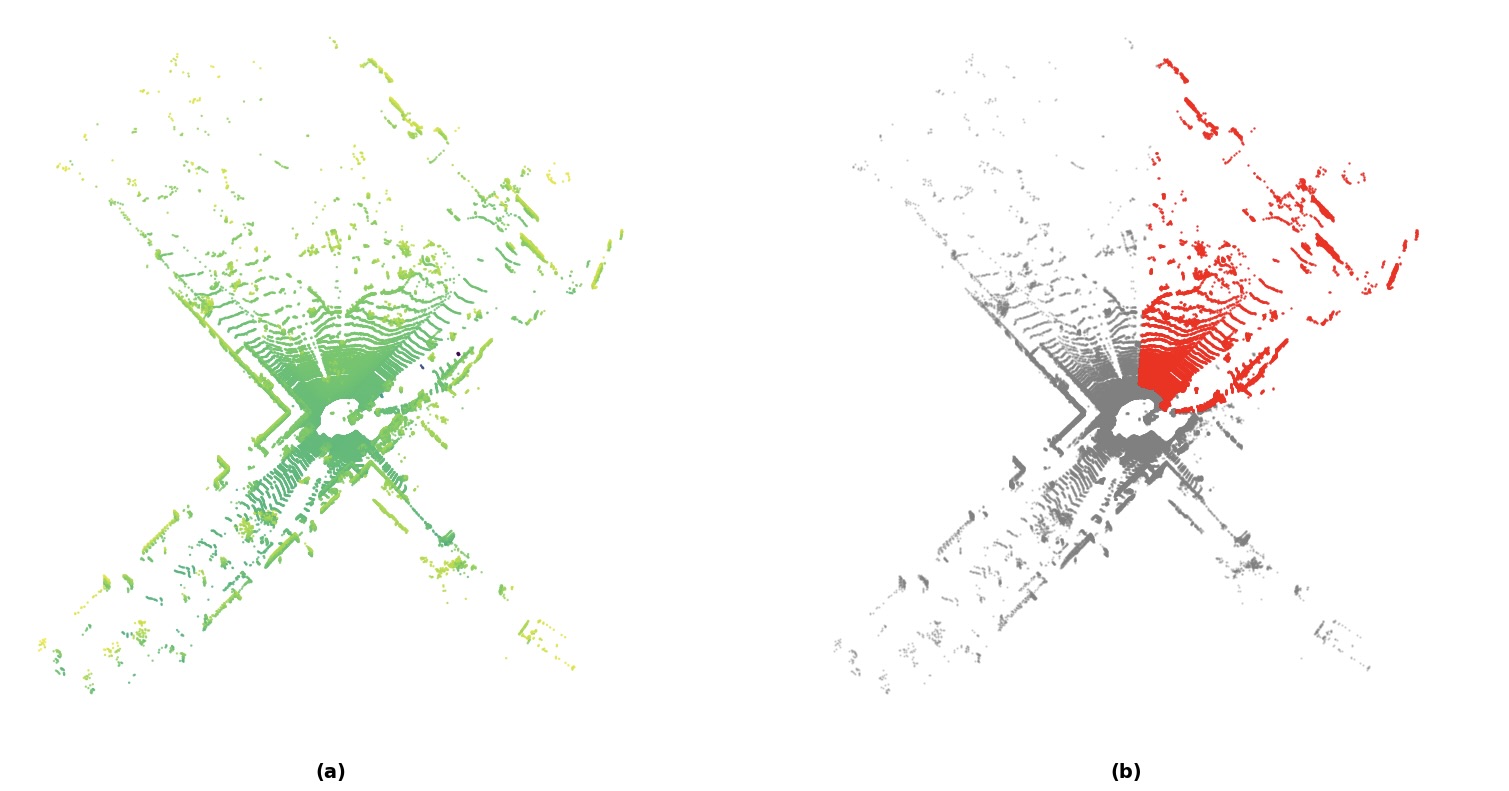}
    \vspace{-0.2cm}
    {\footnotesize (a) \hspace{4.5cm} (b)} \vspace{-0.05cm}
    \caption{Illustration of LiDAR point cloud and camera FOV. (a) Raw 360° LiDAR point cloud. (b) Camera FOV from the same scene; the \textcolor{red}{red region} indicates the portion of the LiDAR points visible from the camera perspective.}\vspace{-0.2cm}
    \label{fig:topview}
\end{figure}

Despite promising accuracy gains, existing fusion approaches face three key issues that hinder practical deployment.
(i) Field-of-View (FOV) mismatch: Differences in sensing ranges between LiDAR and cameras create substantial coverage gaps (see Fig.~\ref{fig:topview}). A significant portion of LiDAR points fall outside the camera’s field of view, limiting cross-modal correspondence. For instance, approximately $23.2\%$ and $33.6\%$ of LiDAR points are not covered by the camera FOV in the nuScenes and Waymo datasets, respectively~\cite{caesar2020nuscenes, sun2020scalability}.
Most fusion methods discard these out-of-FOV points during training and inference \cite{zhuang2021perception, huang2020epnet}. Discarding valuable geometric information will weaken generalization performance.
(ii) Computational overhead: It poses a critical challenge for real-time systems; multi-modal inference must process both images and point clouds, introducing substantial latency and memory cost due to heavy fusion modules and transformer-based designs~\cite{li2023mseg3d, zhuang2021perception}.
(iii) Limited utilization of Vision Foundation Models (VFMs): While 2D vision has been transformed by large-scale pretraining \cite{oquab2024dinov2}, 3D semantic segmentation still largely relies on 3D-only architectures trained from scratch \cite{wu2024point, zhou2021cylinder3d}.
Recent attempts at 2D-to-3D transfer either rely on implicit distillation and suffer from modality gaps \cite{yan2022_2dpass}, or require extracting heavy 2D features at inference time, increasing latency and memory consumption \cite{abouzeid2025dino}.

To address these limitations, we propose \texttt{xModel-KD}, a lightweight framework that explicitly transfers rich 2D semantic knowledge from a frozen 2D teacher into a compact 3D network during training only, incurring zero inference overhead. The central idea is to decouple cross-modal learning from deployment: rather than fusing 2D features at inference, we treat the 2D model as a strong teacher and train the 3D backbone to internalize its semantic priors.

A natural question then arises: \textit{how can 2D semantic knowledge be effectively transferred to the 3D model?} To this end, \texttt{xModel-KD} adopts a multi-scale contrastive distillation strategy. Hierarchical representations from intermediate and deep layers of the frozen 2D backbone are aligned with multi-resolution features from the 3D backbone. Lightweight projection heads map both modalities into a shared embedding space, where a contrastive objective pulls matched point–pixel pairs (via LiDAR-to-image projection) closer while pushing non-matching pairs apart.

Unlike single-layer distillation, multi-scale alignment allows the 3D network to absorb complementary 2D priors across different levels of abstraction: intermediate layers convey spatial structure and boundary cues, while deeper layers provide high-level semantic discrimination. This enriched feature hierarchy also benefits out-of-FOV points. Even without direct pixel correspondences, these points inherit stronger representations through shared 3D parameters learned under 2D supervision.
At deployment, the 2D branch and all cross-modal components are removed, retaining only the enhanced 3D backbone. Consequently, \texttt{xModel-KD} delivers VFM-level semantic priors without requiring images, fusion modules, or additional memory and latency at runtime.

Our main contributions are summarized as follows:
\begin{itemize}
    \item Proposing \texttt{xModel-KD}, a training-only cross-modal knowledge distillation framework that transfers semantic knowledge from frozen 2D teachers into 3D segmentation networks with {zero inference overhead}.
    \item Introducing a multi-scale contrastive objective to align hierarchical 2D and 3D representations, enabling effective transfer of both intermediate (boundary/structure) and deep (semantic) priors.
    \item Designing a lightweight and deployable learning pipeline that mitigates FOV mismatch while preserving inference efficiency by eliminating the 2D pathway at test time.
\end{itemize}

\section{Related Work}
Semantic segmentation for 3D scene understanding has witnessed rapid progress over the past decade, driven by advances in deep learning and the increasing availability of large-scale LiDAR and image datasets. Existing approaches can be broadly categorized into point-based, projection-based, voxel-based, camera-based, and multi-modal learning frameworks. Each paradigm offers distinct advantages in terms of representation power, computational efficiency, and robustness, while also exhibiting inherent limitations. In this section, we review representative works across these categories and discuss their contributions and shortcomings, particularly in the context of LiDAR semantic segmentation and cross-modal learning.

\subsection{Point-based Methods}

These methods aim to approximate a permutation-invariant set function through per-point Multi-Layer Perceptrons (MLPs). PointNet~\cite{qi2017pointnet} pioneered this line of research, and subsequent studies have introduced point-wise MLPs \cite{qi2017pointnetpp,wang2019dgcnn}, adaptive weighting mechanisms, and pseudo-grid strategies to capture local geometric structures, while others have explored non-local operators to model long-range dependencies. However, these methods often suffer from inefficiency in LiDAR applications due to the computational cost of point sampling and grouping operations.

\subsection{Projection-based Methods}\label{AA}

This category of models provides computational efficiency by projecting point clouds onto 2D representations, enabling the direct application of conventional CNNs. Common projection strategies include planar \cite{lawin2017deep,wu2019pointconv}, spherical, and hybrid projections. Nonetheless, projection inevitably leads to information loss, and these methods have reached a performance bottleneck in segmentation accuracy.

\subsection{Voxel-based Methods}\label{AB}

Voxel-based approaches have recently become dominant for balancing accuracy and efficiency. Sparse Convolution~\cite{tang2020spvconv} techniques improve upon traditional 3D CNNs by storing only non-empty voxels in hash tables and performing convolutions selectively, thus enhancing computational efficiency. Building on this idea, several advanced architectures have been proposed. Cylinder3D \cite{zhou2021cylinder3d} introduces cylindrical voxels with an asymmetric network design to boost performance.

\subsection{Camera-based Methods}\label{AC}

Camera-based semantic segmentation focuses on assigning pixel-level semantic labels to 2D images. The Fully Convolutional Network (FCN)~\cite{wu2019pointconv} was the pioneering framework in this field, introducing an end-to-end convolutional architecture derived from image classification networks. Subsequent research has substantially advanced performance by incorporating multi-scale feature learning strategies~\cite{lin2016piecewise}, and attention mechanisms \cite{huang2019ccnet}. These approaches have significantly enhanced contextual understanding and spatial consistency in segmentation results. However, camera-only methods are inherently limited by factors such as lighting conditions, occlusions, and the absence of direct depth information, which restrict their robustness in complex real-world environments.

\subsection{Multi-modal Learning}\label{AD}

This approach has emerged as a powerful paradigm %for enhancing 3D scene understanding 
by jointly exploiting the complementary strengths of RGB images and LiDAR point clouds. Early fusion-based approaches project LiDAR points onto image planes to establish point-to-pixel correspondences and inject visual features into point representations. While effective, such methods are constrained by FOV mismatch and the requirement for tightly aligned multi-modal data at inference time.

\begin{figure*}[!htp]
    \centering
    \begin{tikzpicture}[node distance=1cm]

% Four-equation block

% % Four-equation block
% \node[draw, rounded corners, fill=gray!10,
%       inner sep=6pt, align=left,
%       anchor=north west] (eq_block) at (0,0) {
%     $\displaystyle O_{2D} \in \mathbb{R}^{b \times H \times W \times K}$\\[4pt]
%     $\displaystyle I \in \mathbb{R}^{b \times H \times W \times 3}$\\[4pt]
%     $\displaystyle O \in \mathbb{R}^{b \times p \times K}$\\[4pt]
%     $\displaystyle \mathcal{L}_{total}
%     = \lambda_{3D}\mathcal{L}_{3D\_seg}
%     + \lambda_{2D}\mathcal{L}_{2D\_seg}
%     + \lambda_{contrast}\mathcal{L}_{contrast}$
% };
\node  (img_lidar) at (-8, 8) {\includegraphics[width=4cm, height=1.5cm]{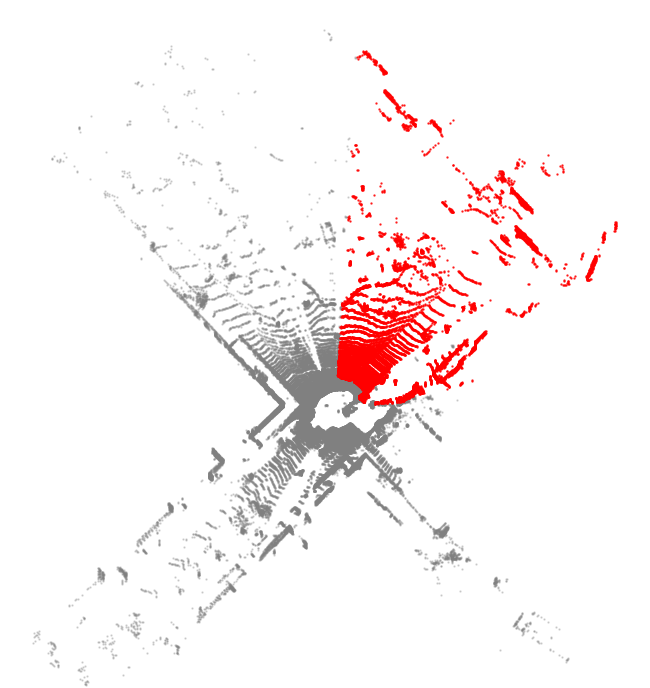}};
\node[font=\footnotesize, text=gray, anchor=south] 
    at ([xshift=0cm, yshift=-0.3cm]img_lidar.north) {Sample of a LiDAR Input};
\node (img_rgb)   at (-8, 6) {\includegraphics[width=4cm, height=1.4cm]{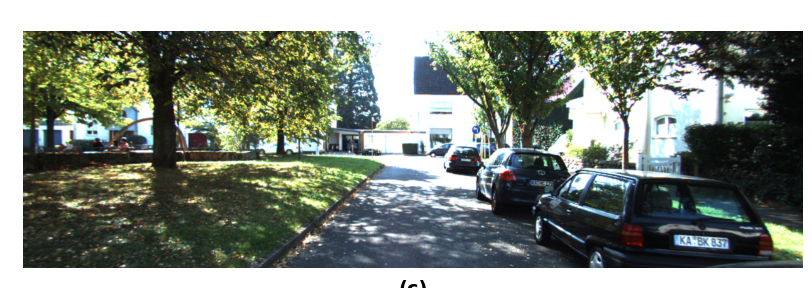}};
\node[font=\footnotesize, text=gray, anchor=south] 
    at ([xshift=0cm, yshift=-0.3cm] img_rgb.north) {Sample of a RGB Input };
\node (label_lidar) at (-8, 4) {\includegraphics[width=4cm, height=1.5cm]{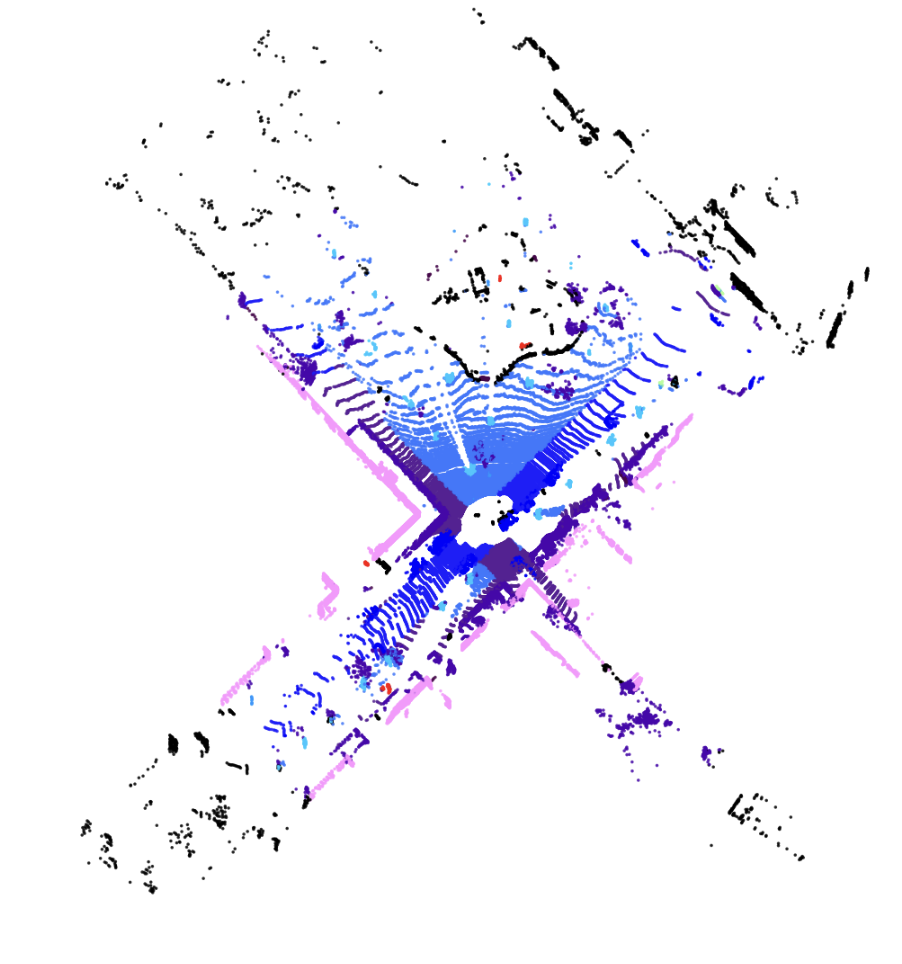}};
\node[font=\footnotesize, text=gray, anchor=south] 
    at ([xshift=0cm, yshift=-0.2cm] label_lidar.north) {The Point Cloud Segmentation wrt the Input};
\node (label_rgb)   at (-8, 2) {\includegraphics[width=4cm, height=1.4cm]{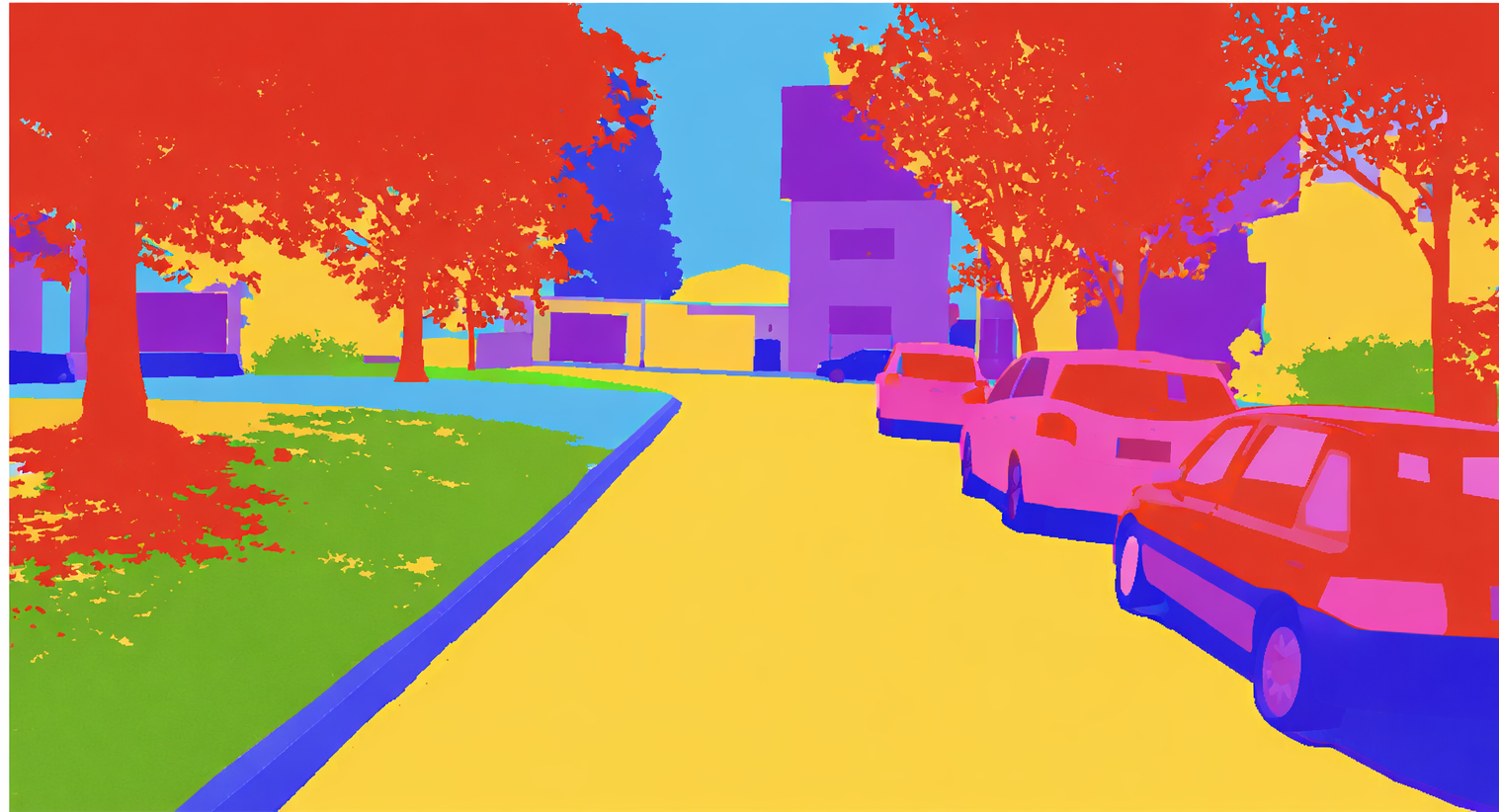}};
\node[font=\footnotesize, text=gray, anchor=south] 
    at ([xshift=0cm, yshift=-0.2cm] label_rgb.north) {The Segmenation of the RGB Counterpart};
\node[inputbox, minimum width=3cm, text width=2cm, inner sep=0pt] (input_lidar) at (-4, 7.8) {LiDAR Point Cloud};
\node[inputbox,minimum width=3cm, text width=2cm, inner sep=0pt] (input_rgb) at (-4, 5.8) {RGB Image};

% Backbone networks
\node[backbone,minimum width=2.5cm] (backbone_3d) [right=of input_lidar] {\shortstack{3D Backbone\\(SPVCNN)}};
\node[backbone,minimum width=2.5cm] (backbone_2d) [right=of input_rgb] {\shortstack{2D Backbone\\(ResNet50)}};

\node[annotation] (feat_3d) 
    at ([xshift=0.2cm, yshift=0.2cm]backbone_3d.east) 
    {$\mathbf{f}_{3D}$};

\node[annotation] (feat_2d) at ([xshift=0.2cm, yshift=0.2cm]backbone_2d.east) {$\mathbf{f}_{2D}$};

% Projection heads
\node[loss,minimum width=2cm, text width=2cm, inner sep=0pt] (proj_3d) [right=1cm of backbone_3d] {Proj. $\mathbf{Z}_{3D}$};
\node[loss,minimum width=2cm, text width=2cm, inner sep=0pt] (proj_2d)[right=1cm of backbone_2d] {Proj. $\mathbf{Z}_{2D}$};

\node[module, fill=contrastcolor!40,minimum width=2cm, text width=2cm, inner sep=0pt] (contrastive) [below right=0cm and 0.3cm of proj_2d] {
    \textbf{Contrastive Module}
};

% Projection heads
\node[loss,minimum width=2cm, text width=2cm, inner sep=0pt, fill=yellow!30] (proj_3di) [below =1cm of proj_2d] {Inverse Proj. $\mathbf{Z}_{3D}$};
\node[loss,minimum width=2cm, text width=2cm, inner sep=0pt, fill=yellow!30] (proj_2di)[below =3cm of proj_2d] {Inverse Proj. $\mathbf{Z}_{2D}$};

% Projection heads
\node[backbone,minimum width=2.5cm, fill=orange!15] (3D_dec) [below =1cm of backbone_2d] {3D Decoder };
\node[backbone,minimum width=2.5cm, fill=orange!15] (2d_dec)[below =3cm of backbone_2d] {2D Decoder };

% Projection heads
\node[output,minimum width=3cm, text width=2cm, inner sep=0pt] (3D_out) [left=  of 3D_dec] {3D Segmentation };
\node[output,minimum width=3cm, text width=2cm, inner sep=0pt] (2d_out)[left= of 2d_dec] {2D Segmentation };

\node[rounded corners,
      inner sep=4pt,
      align=center,
      anchor=south] (eq_block)
      at ([xshift=0.5cm,yshift=-0.4mm]2d_out) {
    {\scriptsize $\displaystyle O_{2D} \in \mathbb{R}^{b \times H \times W \times K}$}\\[3pt]
};

\node[rounded corners,
      inner sep=4pt,
      align=center,
      anchor=south] (eq_block)
      at ([xshift=0.5cm,yshift=-0.4mm]3D_out) {
    {\scriptsize $\displaystyle O_{3D} \in \mathbb{R}^{b \times p \times K}$}\\[4pt]
};

\node[rounded corners,
      inner sep=4pt,
      align=center,
      anchor=south] (eq_block)
      at ([xshift=0.5cm,yshift=-0.4mm]input_rgb) {
    {\scriptsize $\displaystyle I \in \mathbb{R}^{b \times H \times W \times 3}$}\\[4pt]
};

\node[rounded corners,
      inner sep=4pt,
      align=center,
      anchor=south] (eq_block)
      at ([xshift=.5cm,yshift=-0.4mm]input_lidar) {
    {\scriptsize $\displaystyle I \in \mathbb{R}^{b \times 3 \times p}$}\\[4pt]
};

\draw[arrow] (input_lidar) -- (backbone_3d) ;
\draw[arrow] (backbone_3d) -- (proj_3d);
\draw[arrow,dashed] (input_rgb) -- (backbone_2d) ;
\draw[arrow,dashed] (backbone_2d) -- (proj_2d);
\draw[arrow] (proj_3di) -- (3D_dec) ;
\draw[arrow] (3D_dec) -- (3D_out);
\draw[arrow,dashed] (proj_2di) -- (2d_dec) ;
\draw[arrow,dashed] (2d_dec) -- (2d_out);

\draw[->, line, dashed] (proj_2d.east) -| ([xshift=-5mm]contrastive.north);
\draw[->, line] (proj_3d.east) -| ([xshift=5mm]contrastive.north);

\draw[->, line] ([xshift=-5mm]contrastive.south) |- (proj_3di.east);
\draw[->, line, dashed] ([xshift=5mm]contrastive.south) |- (proj_2di.east);

\draw[->, thick, green!70!black]
    ([yshift=-3mm]backbone_3d.east) -- ++(0.5cm,0)  % go right
    |- ++(0,-2.5cm)               % go down
    -| (3D_dec.north); % go left to slightly-left of north

\draw[->, thick, dashed, green!70!black]
    ([yshift=-3mm]backbone_2d.east) -- ++(0.8cm,0)  % go right
    |- ++(0,-2.5cm)               % go down
    -| (2d_dec.north); % go left to slightly-left of north

\begin{pgfonlayer}{background}
  \node[draw=none, fill=Brown!15, rounded corners=6pt, inner xsep=0pt, inner ysep=10pt,
        fit=(proj_3d)(proj_2d)(contrastive)(proj_3di)(proj_2di),
        inner sep=10pt,
        label={[text=Brown, font=\footnotesize\bfseries, anchor=south, yshift=-0.1cm]above:{S\texttimes~Multi-scale Projection Head}}] (multiscale_box) {};
\end{pgfonlayer}

% -------------------- LEGEND --------------------

% --- Small square style ---
\tikzset{
    legendsq/.style={
        draw=black,
        minimum width=0.35cm,
        minimum height=0.35cm,
        inner sep=0pt
    }
}

% --- Legend rows ---
% \node[anchor=west] at ([xshift=-12cm,yshift=-1.7cm]2d_dec.south)
%     {\tikz{\node[legendsq, fill=inputcolor!30,inner sep=0pt]{Input};} \hspace{0pt} Input};

\node[anchor=west] at ([xshift=-10cm,yshift=-1.7cm]2d_dec.south)
    {\tikz{\node[legendsq, fill=inputcolor!30,inner sep=2pt, rounded corners=2]{\footnotesize Input};}};

% \node[anchor=west] at ([xshift=-10.2cm,yshift=-1.7cm]2d_dec.south)
%     {\tikz{\node[legendsq, fill=backbonecolor!30]{};} \hspace{0pt} Encoder};

\node[anchor=west] at ([xshift=-8.5cm,yshift=-1.7cm]2d_dec.south)
    {\tikz{\node[legendsq, fill=backbonecolor!30 ,inner sep=2pt, rounded corners=2]{\footnotesize Encoder};}};

% \node[anchor=west] at ([xshift=-7.9cm,yshift=-1.7cm]2d_dec.south)
%     {\tikz{\node[legendsq, fill=contrastcolor!30]{};} \hspace{0pt} Contrastive
% Module};

\node[anchor=west] at ([xshift=-3.5cm,yshift=-1.7cm]2d_dec.south)
    {\tikz{\node[legendsq, fill=contrastcolor!30,inner sep=2pt, rounded corners=2]{\footnotesize Contrastive Module};}};

\node[anchor=west] at ([xshift=-6.5cm,yshift=-1.7cm]2d_dec.south)
    {\tikz{\node[legendsq, fill=losscolor!30 ,inner sep=2pt, rounded corners=2]{ \footnotesize Projection Head};}};

\node[anchor=west] at ([xshift=0cm,yshift=-1.7cm]2d_dec.south)
    {\tikz{\node[legendsq, fill=yellow!30,inner sep=2pt, rounded corners=2]{\footnotesize Inverse Projection};} };

\node[anchor=west] at ([xshift=3cm,yshift=-1.7cm]2d_dec.south)
    {\tikz{\node[legendsq, fill=orange!15 ,inner sep=2pt, rounded corners=2]{\footnotesize Decoder};}};

\node[anchor=west] at ([xshift=5cm,yshift=-1.7cm]2d_dec.south)
    {\tikz{\node[legendsq, fill=outputcolor!30,inner sep=2pt, rounded corners=2]{\footnotesize Output};}};

% ---- Arrows ----
\node[anchor=west] (l6) at ([xshift=-5cm,yshift=-1.5cm]2d_out.west)
    {\begin{tikzpicture}
        \draw[arrow] (-3,0) -- (-1.5,0);
        \node[right] at (-1.5,0) {: Inference + Training};
     \end{tikzpicture}};

\node[anchor=west] (l7) at ([xshift=0.5cm,yshift=-1.5cm]2d_out.west)
    {\begin{tikzpicture}
        \draw[arrow,dashed] (-2.5,0) -- (-1,0);
        \node[right] at (-1,0) {,};
        \draw[->, thick, dashed, green!70!black] (-0.6,0) -- (0.9,0);
        \node[right] at (0.9,0) {: Training Only};
     \end{tikzpicture}};

\node[anchor=west] (l8) at ([xshift=7.0cm,yshift=-1.5cm]2d_out.west)
    {\begin{tikzpicture}
        \draw[->, thick, green!70!black] (0,0) -- (1.5,0);
        \node[right] at (1.5,0) {: Auxiliary/Skip Flow};
     \end{tikzpicture}};
% --------------------------------------------------

% \node[anchor=south east, align=left] (loss_eq) at ($(current bounding box.south east)+(0.7cm,0.5cm)$) {
%     $\mathcal{L}_{3D\_seg}$ (3D Segmentation Loss)\\[4pt]
%     $\mathcal{L}_{2D\_seg}$ (2D Segmentation Loss)\\[4pt]
%     $\mathcal{L}_{contrast}$ (Contrastive Loss): \\[4pt]

% };

\end{tikzpicture}
    \caption[Unified 2D-3D semantic understanding framework]{Overview of the proposed \texttt{xModel-KD}. It jointly learns unified representations from LiDAR and RGB data using separate 3D  and 2D encoders. Multi-scale features are projected into a shared embedding space and aligned via a contrastive module. The aligned features are fed into 2D and 3D decoders for simultaneous segmentation. Cross-modal distillation enforces prediction consistency during training, while only the 3D stream is used at inference.}
    \label{fig:framework}
\end{figure*}
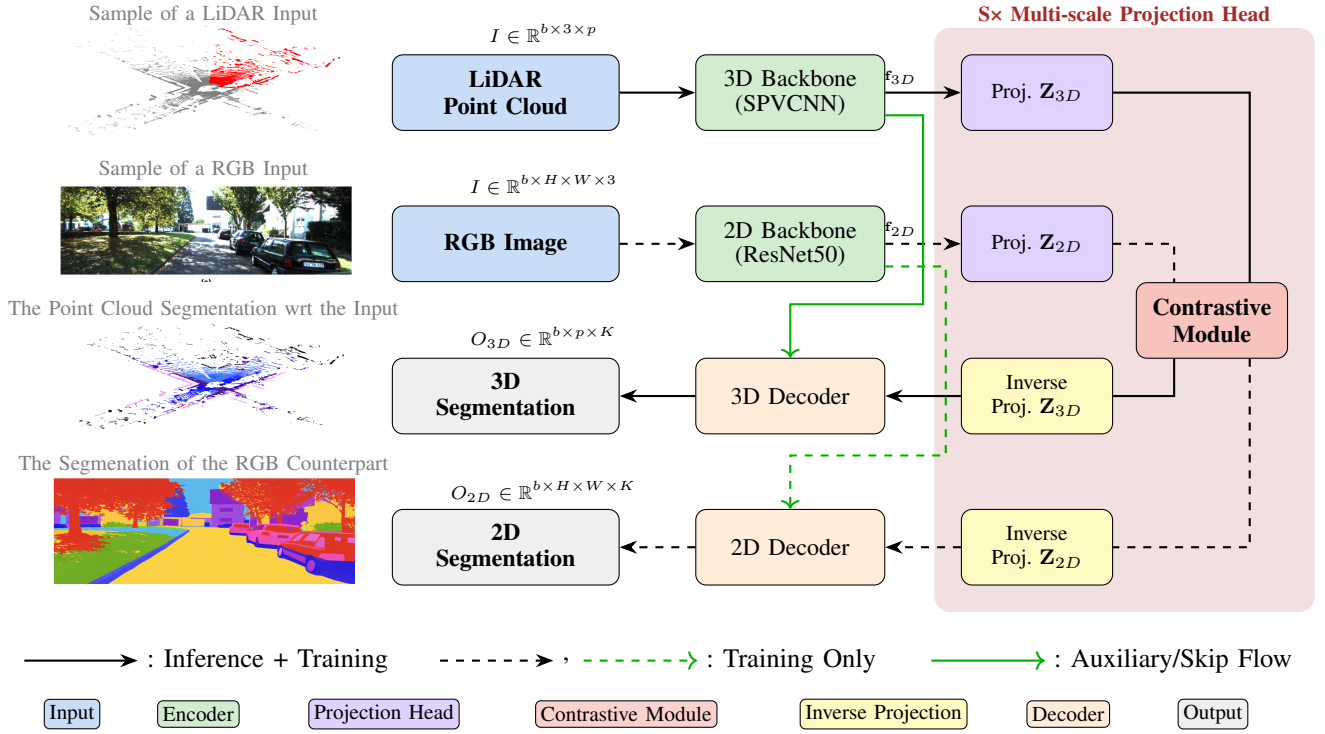

To alleviate these limitations, Yan~\textit{et al.}~\cite{yan2022_2dpass} propose {2DPASS}, a training-only multi-modal distillation framework that leverages camera images as semantic priors to enhance LiDAR representations. By employing a multi-scale fusion-to-single knowledge distillation strategy, 2DPASS transfers rich 2D semantic information into a pure 3D network during training, enabling image-free inference. % while achieving state-of-the-art performance on SemanticKITTI and nuScenes.
More recently, foundation-model-based approaches have been explored to further strengthen cross-modal learning. Knaebel~\textit{et al.}~\cite{abouzeid2025dino} introduce DINO In The Room (DITR), which injects features extracted from large-scale 2D vision foundation models into 3D segmentation backbones via 2D-to-3D feature projection. In addition, they propose a distillation variant that aligns 3D representations with frozen 2D foundation features, allowing the model to retain strong semantic priors even when images are unavailable at inference. This approach demonstrates consistent improvements across both indoor and outdoor benchmarks.
Addressing robustness and modality-misalignment issues, Sun~\textit{et al.}~\cite{sun2024unitomulti} propose a Uni-to-Multi Modal Knowledge Distillation framework for bidirectional LiDAR-camera semantic segmentation. Their method introduces a bidirectional feature fusion and imputation mechanism to handle out-of-FOV points and missing image features, combined with a unimodal teacher to a multi-modal student distillation strategy. %This design effectively mitigates augmentation misalignment and improves robustness under sensor degradation, achieving strong gains on nuScenes, Waymo, and SemanticKITTI.

Overall, these works highlight a growing trend toward leveraging cross-modal knowledge transfer and distillation to overcome the inherent limitations of direct fusion. %, enabling robust and efficient multi-modal 3D perception.

\section{Methodology}

%This section presents our {Cross-Modal Feature Fusion Framework} for joint 2D-3D semantic understanding combining LiDAR point clouds and RGB images. 

The proposed approach, as illustrated in Fig.~\ref{fig:framework} subsumes three sub-tasks: (i) multi-scale feature extraction with separate 2D and 3D encoder-decoder (EnDec) backbones, (ii) multi-scale contrastive alignment for cross-modal feature fusion, and (iii) cross-modal knowledge distillation.

\subsection{Multi-scale Feature Extraction }

Table~\ref{tab:architecture} provides a layer-by-layer detail of the proposed \texttt{xModel-KD}, which processes multi-scale features hierarchically from both RGB and LiDAR inputs. 
The LiDAR input $\mathbf{X}\!\in\!\mathbb{R}^{[b,p,4]}$ is processed by a 3D backbone, producing point-wise embeddings $\mathbf{F}_{3D}\!\in\!\mathbb{R}^{[b,p,l_{c}]}$, where each point is mapped to a $l_{c}$-dimensional latent feature space.
In parallel, the RGB image 
$\mathbf{I}\!\in\!\mathbb{R}^{[b,3,H,W]}$ is encoded using a 2D backbone, 
yielding spatial feature maps
$\mathbf{F}_{2D}\!\in\!\mathbb{R}^{[b,H',W',l_{h}]}$, where $l_{h}$ denotes the number of feature channels. 
To enable cross-modal alignment, the backbone features from both modalities are mapped into a shared embedding space through modality-specific projection heads. 
Specifically, the 3D features, $\mathbf{F}_{3D}\!\in\!\mathbb{R}^{[b,p,l_{c}]}$ are projected to 
$\mathbf{Z}_{3D}\!\in\!\mathbb{R}^{[b,p,d_p]}$, 
while the 2D features $\mathbf{F}_{2D}\!\in\!\mathbb{R}^{[b,H',W',l_{h}]}$ are projected to 
$\mathbf{Z}_{2D}\!\in\!\mathbb{R}^{[b,H',W',d_p]}$, 
where $d_p$ denotes the projection dimension (set to 128 in this case). 
Thus, both the modalities are projected into the same dimensional embedding space ($d_p$), ensuring that $\mathbf{Z}_{3D}$ and $\mathbf{Z}_{2D}$ have matching feature dimensions at each scale, which is required for contrastive alignment.
Mapping both modalities into a shared $d_p$-dimensional space facilitates contrastive learning and cross-modal feature alignment. After cross-modal alignment in the shared embedding space, the projected features are mapped back to their respective modality-specific feature dimensions through inverse projection layers. 
Specifically, the aligned 3D embeddings 
$\mathbf{Z}_{3D} \in \mathbb{R}^{[b,p,d_p]}$ 
are transformed into 
$\tilde{\mathbf{F}}_{3D} \in \mathbb{R}^{[b,p,l_{c}]}$, 
where the inverse projection restores the feature dimensionality required by the subsequent 3D decoder. 
Similarly, the aligned 2D embeddings 
$\mathbf{Z}_{2D} \in \mathbb{R}^{[b,H',W',d_p]}$ 
are mapped to 
$\tilde{\mathbf{F}}_{2D} \in \mathbb{R}^{[b,H',W',l_{h}]}$ 
during training. 
These inverse transformations allow the modality-specific decoders to operate in their corresponding feature spaces while preserving the cross-modal alignment learned in the projection space.

\begin{table}[!tp]
\centering
\caption{Layer specifications of the proposed \texttt{xModal-KD}.}
\label{tab:architecture}
%\small
\setlength{\tabcolsep}{4pt}
\renewcommand{\arraystretch}{1.0}
\begin{tabularx}{\columnwidth}{l>{\centering\arraybackslash}X>{\centering\arraybackslash}X}
\hline
\textbf{Component} & \textbf{Input} & \textbf{Output} \\
\hline\hline
\rowcolor{Green!20}
\multicolumn{3}{c}{{Encoders}} \\
\hline
LiDAR input  - $\mathbf{X}$ & $[b,p,4]$ & -- \\
RGB input  - $\mathbf{I}$ & $[b,3,H,W]$ & -- \\
3D Backbone (SPVCNN) & $\mathbf{X}$ & $[b,p,l_{c}]$ \\
2D Backbone (ResNet50) & $\mathbf{I}$ & $[b,H',W',l_{h}]$ \\
\hline
\rowcolor{Purple!20}
\multicolumn{3}{c}{{Projection Head}} \\
\hline
Proj. Head - $\mathbf{Z}_{3D}$ & $l_{c}$ & $\mathbf{d}_{p}$ - 128 \\
Proj. Head - $\mathbf{Z}_{2D}$ (train only) & $l_{h}$ & $\mathbf{d}_{p}$ - 128 \\
Contrastive Module (train only) & $(\mathbf{Z}_{3D},\mathbf{Z}_{2D})$ & $\mathcal{L}_{contrast}$ \\
\hline
\rowcolor{Yellow!20}
\multicolumn{3}{c}{{Inverse Projection}} \\
\hline
Inverse Proj. - $\mathbf{Z}_{3D}$ & $\mathbf{d}_{p}$ - 128 & $l_{h}$ \\
Inverse Proj. - $\mathbf{Z}_{2D}$ (train only) & $\mathbf{d}_{p}$ - 128 & $l_{c}$ \\
\hline
\rowcolor{orange!15}
\multicolumn{3}{c}{{Decoders and Outputs}} \\
\hline
3D Decoder & $[b,p,l_{h}]$ & $[b,p,K]$ \\
%3D Seg. Head & $d_h$ & $[b,p,K]$ \\
2D Decoder (train only) & $[b,H',W',l_{c}]$ & $[b,H,W,K]$ \\
%2D Seg. Head (train only) & $d_h$ & $[b,H',W',K]$ \\
\hline \hline
\multicolumn{3}{p{0.95\columnwidth}}{
\footnotesize Note: $b$ - batch size; $p$ - number of data points in each LiDAR point cloud sample; $H, W$ - spatial dimension of the inpput RGB image. }
\end{tabularx}
\end{table}

\begin{algorithm}[tp!]
\caption{Multi-Scale Contrastive Alignment}
\label{alg:contrastive}
\begin{algorithmic}[1]
\Require $\{f_{3D}^{s}\}_{s=1}^{S}$, $\{f_{2D}^{s}\}_{s=1}^{S}$ (multi-scale features), $\text{Proj}_{3D}^{s}$, $\text{Proj}_{2D}^{s}$ (projection heads), $\tau$ (temperature)
\Ensure $\mathcal{L}_{\text{contrast}}$ (total contrastive loss)

\State $\mathcal{L}_{\text{total}} \gets 0$

\For{$s = 1$ to $S$}
    \State $\mathbf{z}_{3D}^{s} \gets \text{Proj}_{3D}^{s}(f_{3D}^{s})$ \textcolor{gray}{\footnotesize \Comment{Project 3D features}}
    \State $\mathbf{z}_{2D}^{s} \gets \text{Proj}_{2D}^{s}(f_{2D}^{s})$ \textcolor{gray}{\Comment{\footnotesize \footnotesize Project 2D features}}
    
    \State $\mathbf{z}_{3D}^{s} \gets \text{Normalize}(\mathbf{z}_{3D}^{s})$ \textcolor{gray}{\Comment{\footnotesize L2 normalization}}
    \State $\mathbf{z}_{2D}^{s} \gets \text{Normalize}(\mathbf{z}_{2D}^{s})$
    
    \State $\text{logits} \gets \frac{1}{\tau} \mathbf{z}_{3D}^{s} \cdot (\mathbf{z}_{2D}^{s})^{\top}$ \textcolor{gray}{\Comment{\footnotesize Compute similarity matrix}}
    
    \State $\text{labels} \gets \text{arange}(N)$ \textcolor{gray}{\Comment{\footnotesize Positive pairs on diagonal}}
    
    \State $\mathcal{L}_{\text{contrast}}^{s} \gets \text{CrossEntropy}(\text{logits}, \text{labels})$ \textcolor{gray}{\Comment{\footnotesize NT-Xent loss}}
    
    \State $\mathcal{L}_{\text{total}} \gets \mathcal{L}_{\text{total}} + \frac{1}{S} \mathcal{L}_{\text{contrast}}^{s}$
\EndFor

\State \Return $\mathcal{L}_{\text{total}}$
\end{algorithmic}
\end{algorithm}

\subsection{Multi-Scale Contrastive Alignment}

To establish cross-modal correspondences, as summarized in Algorithm~\ref{alg:contrastive} also As illustrated in Fig.~\ref{fig:framework}, the dashed box denotes the multi-scale contrastive module, which is repeated across S scales to align 2D and 3D features, we employ a multi-scale contrastive learning module that aligns 2D and 3D features in a shared embedding space.
At each scale $s$, we extract features $f_{3D}^{s}$ from the 3D backbone and $f_{2D}^{s}$ from the 2D backbone. These features are projected into a common embedding space using modality-specific projection heads, as defined in \eqref{eq-proj}.
\begin{equation}
\mathbf{z}_{3D}^{s} = \text{Proj.}_{3D}^{s}(f_{3D}^{s}), \quad \mathbf{z}_{2D}^{s} = \text{Proj.}_{2D}^{s}(f_{2D}^{s})\label{eq-proj}
\end{equation}
Using known camera-to-LiDAR calibration parameters, we establish point-to-pixel correspondence relationships. A normalized temperature-scaled cross-entropy (NT-Xent) contrastive loss given by \eqref{eq-contras}  encourages matched pairs to have similar embeddings.
\begin{equation}
\mathcal{L}_{\text{contrast}}^{s} = - \log 
\frac{\exp(\text{sim}(\mathbf{z}_{3D}^{s}, \mathbf{z}_{2D}^{s}) / \tau)}
{\sum_{j=1}^{N} \exp(\text{sim}(\mathbf{z}_{3D}^{s}, \mathbf{z}_{2D}^{(j)}) / \tau)}, \label{eq-contras}
\end{equation}
where $\text{sim}(a,b) = \frac{a \cdot b}{\|a\|\|b\|}$ is cosine similarity, $\tau$ is the temperature parameter, and $N$ is the batch size. The total contrastive loss aggregates across scales is formulated as \eqref{eq-contra}

\begin{equation}
\mathcal{L}_{\text{contrast}} = \frac{1}{S} \sum_{s=1}^{S} \mathcal{L}_{\text{contrast}}^{s}\label{eq-contra}\
\end{equation}
Through this contrastive objective, both modalities learn mutually aligned representations that capture complementary geometric and semantic information.

\subsection{Cross-modal Knowledge Distillation}

To maintain prediction consistency and encourage knowledge transfer between modalities, we apply the KL divergence between matched 2D pixel predictions and 3D point predictions using established point–pixel correspondences. Let $\Omega$ denote the set of LiDAR points that fall within the camera field of view. Using known camera–LiDAR calibration parameters, each point $i\!\in\!\Omega$ is projected to its corresponding pixel coordinate $\pi(i)$ on the image plane.
Thus, the knowledge distillation loss, $\mathcal{L}_{\text{KD}}$, can be calculated as in \eqref{eq-Kl}. 
\begin{equation}
\mathcal{L}_{\text{KD}} =
\frac{1}{|\Omega|}
\sum_{i \in \Omega}
\text{KL}\big(
P_{2D}(\pi(i)) \,\|\, P_{3D}(i)
\big),\label{eq-Kl}\
\end{equation} %Will you include qualitative results from test set?
where $P_{3D}(i)$ denotes the predicted class distribution for LiDAR point $i$, computed using $\text{softmax}(\hat{y}_{3D}(i))$ and $P_{2D}(\pi(i))$ denotes the predicted distribution at the corresponding image pixel, estimated using $\text{softmax}(\hat{y}_{2D}(\pi(i)))$.

The KL divergence is, therefore, computed only over matched point–pixel pairs. Points outside the camera field of view are excluded from this loss but indirectly benefit from shared 3D parameters optimized under matched supervision.
The combination of multi-scale contrastive alignment, weighted fusion, and multi-task optimization enables the framework to learn rich, modality-aligned representations that leverage complementary geometric and semantic cues from both LiDAR and camera data, significantly improving downstream 3D semantic segmentation performance.

% -------------------------
% FULL-WIDTH TABLE: DATASET STATS (SemanticKITTI only)
% -------------------------
\begin{table*}[ht!]
\centering
\caption{Dataset statistics and evaluation protocol}
\label{tab:dataset_stats}
\renewcommand{\arraystretch}{1.15}
\setlength{\tabcolsep}{6pt}
\begin{tabular}{lcccccc}
\toprule
\textbf{Dataset} & \textbf{Domain} & \textbf{Sensors} & \textbf{\#Classes} & \textbf{Train Split} & \textbf{Val Split} & \textbf{Test} \\
\midrule
SemanticKITTI~\cite{Behley2019SemanticKITTI} & Outdoor & LiDAR + front RGB & 19 &
Seq. 00--07,09--10 (19,130) & Seq. 08 (4,071) & Online (11--21) \\
\bottomrule
\end{tabular}
\end{table*}

\begin{table*}[tp!]
\centering
\caption{A comparative analysis of various semantic models on {SemanticKITTI} test benchmark. Note: baseline is a LiDAR-only model\textbf{Bold} and \underline{underlined} values indicate the best and second-best performance, respectively.}

\label{tab:semantickitti}
%\scriptsize
\setlength{\tabcolsep}{1.3pt}
\renewcommand{\arraystretch}{1}
\begin{tabular}{lcccccccccccccccccccc}
\toprule

\multirow{2}{*}{\textbf{Method}} & \textbf{Overall} & \multicolumn{19}{c}{\textbf{Class-wise mIoU ($\%$)}}\\
 &  \textbf{mIoU } &  \texttt{\scriptsize road} & \texttt{\scriptsize side} & \texttt{\scriptsize park} & \texttt{\scriptsize oth-g} & \texttt{\scriptsize build} & \texttt{\scriptsize car} & \texttt{\scriptsize truck} & \texttt{\scriptsize bike} & \texttt{\scriptsize m-bike} &
\texttt{\scriptsize oth-v} & \texttt{\scriptsize veg} & \texttt{\scriptsize trunk} & \texttt{\scriptsize terr} & \texttt{\scriptsize pers} & \texttt{\scriptsize bicy} & \texttt{\scriptsize motor} & \texttt{\scriptsize fence} & \texttt{\scriptsize pole} & \texttt{\scriptsize sign} \\
\midrule
SqueezeSegV2~\cite{wu2019squeezesegv2} & 39.7 & 88.6 & 67.6 & 45.8 & 17.7 & 73.7 & 81.8 & 13.4 & 18.5 & 17.9 & 14.0 & 71.8 & 35.8 & 60.2 & 20.1 & 25.1 & 3.9 & 41.1 & 20.2 & 26.3 \\
DarkNet53Seg~\cite{Behley2019SemanticKITTI} & 49.9 & 91.8 & 74.6 & 64.8 & 27.9 & 84.1 & 86.4 & 25.5 & 24.5 & 32.7 & 22.6 & 78.3 & 50.1 & 64.0 & 36.2 & 33.6 & 4.7 & 55.0 & 38.9 & 52.2 \\
RangeNet53++~\cite{Milioto2019RangeNetPP} & 52.2 & 91.8 & 75.2 & 65.0 & 27.8 & 87.4 & 91.4 & 25.7 & 25.7 & 34.4 & 23.0 & 80.5 & 55.1 & 64.6 & 38.3 & 38.8 & 4.8 & 58.6 & 47.9 & 55.9 \\
RandLA-Net~\cite{Hu2020RandLANet} & 55.9 & 90.5 & 74.0 & 61.8 & 24.5 & 89.7 & 94.2 & 43.9 & 29.8 & 32.2 & 39.1 & 83.8 & 63.6 & 68.6 & 48.4 & 47.4 & 9.4 & 60.4 & 51.0 & 50.7 \\
KPConv~\cite{Thomas2019KPConv} & 58.8 & 90.3 & 72.7 & 61.3 & 31.5 & 90.5 & 95.0 & 33.4 & 30.2 & 42.5 & 44.3 & 84.8 & 69.2 & 69.1 & 61.5 & 61.6 & 11.8 & 64.2 & 56.4 & 47.4 \\
Cylinder3D~\cite{zhu2021cylinder3d} & 68.9 & \underline{92.2} & 77.0 & 65.0 & 32.3 & 90.7 & 97.1 & 50.8 & 67.6 & 63.8 & 58.5 & 85.6 & 72.5 & 69.8 & 73.7 & 69.2 & 48.0 & 66.5 & 62.4 & 66.2 \\
RPVNet~\cite{Xu2021RPVNet} & \underline{70.3} & \textbf{93.4} & \textbf{80.7} & \underline{70.3} & 33.3 & \underline{93.5} & \textbf{97.6} & 44.2 & \textbf{68.4} & 68.7 & 61.1 & \underline{86.5} & \textbf{75.1} & \underline{71.7} & 75.9 & \underline{74.4} & \underline{43.4} & 72.1 & \underline{64.8} & 61.4 \\
2DPASS~\cite{yan2022_2dpass} & \textbf{72.9} & 89.7 & 74.7 & 67.4 & \underline{40.0} & 93.5 & \underline{97.0} & \underline{61.1} & \underline{63.6} & 63.4 & \underline{61.5} & 86.2 & \underline{73.9} & 71.0 & \textbf{77.9} & \textbf{81.3} & \textbf{74.1} & \textbf{72.9} & \textbf{65.0} & \textbf{70.4} \\
\midrule
Baseline  & 67.0 & 89.8 & 73.8 & 57.0 & \textbf{52.0} & 91.9 & 96.3 & 60.0 & 55.0 & 58.0 & 51.6 & \textbf{86.5} & 65.0 & 71.3 & \underline{76.8} & 55.0 & 55.0 & \underline{68.7} & 60.0 & 49.3 \\
\rowcolor{green!15}
{\texttt{xModel-KD}} & {69.1} & \underline{\textbf{92.9}} & \underline{79.4} & 54.0 & \textbf{50.0} & 89.9 & 96.2 & \textbf{78.5} & 55.0 & \textbf{73.8} & \textbf{55.5} & \textbf{87.9} & 71.4 & \textbf{74.1} & 75.5 & \textbf{59.0} & 55.0 & 57.5 & 58.2 & \textbf{50.4} \\
\bottomrule
\end{tabular}
\end{table*}

\subsection{Loss Function Formulation}

The \texttt{xModel-KD} is trained using a multi-task objective that jointly optimizes segmentation, cross-modal contrastive alignment, and knowledge distillation, as defined in~\eqref{eq-multiloss}.
\begin{equation}
\mathcal{L}_{\text{total}} =
\mathcal{L}_{3D} +
\mathcal{L}_{2D} +
\lambda_{\text{contrast}} \mathcal{L}_{\text{contrast}} +
\lambda_{\text{KD}} \mathcal{L}_{\text{KD}},
\label{eq-multiloss}
\end{equation}
where $\mathcal{L}_{3D}$ and $\mathcal{L}_{2D}$ denote the 3D and 2D segmentation losses, respectively, $\mathcal{L}_{\text{contrast}}$ is the cross-modal contrastive loss weighted by $\lambda_{\text{contrast}}$, and $\mathcal{L}_{\text{KD}}$ is the knowledge distillation loss weighted by $\lambda_{\text{KD}}$.
Hence, the segmentation losses are:
\begin{equation}
\mathcal{L}_{3D} =
\mathcal{L}_{CE}(\hat{y}_{3D}, y_{3D}) +
\mathcal{L}_{\text{Lovász}}(\hat{y}_{3D}, y_{3D}),
\label{eq:l3d}
\end{equation}
\begin{equation}
\mathcal{L}_\text{2D} =
\mathcal{L}_\text{CE}(\hat{y}_{2D}, y_{\text{img}}) +
\mathcal{L}_{\text{Lovász}}(\hat{y}_{2D}, y_{\text{img}}),
\label{eq:l2d}
\end{equation}
where $\hat{y}$ and $y$ denote the predicted and ground-truth labels, respectively. The combination of cross-entropy (CE) and Lovász-Softmax (Lovász) mitigates class imbalance while improving boundary alignment.

\section{Experimental Setup and Performance Analysis}
\label{sec:experiments}

% \subsection{Experimental Setup and Performance Analysis}
% \label{sec:exp_setup}

\subsection{Datasets}

We evaluate \texttt{xModel-KD} on the SemanticKITTI benchmark~\cite{Behley2019SemanticKITTI}, a large-scale outdoor LiDAR semantic segmentation dataset.
SemanticKITTI provides point-wise semantic labels for LiDAR scans captured with a 64-beam sensor.
Following the official split, summarized in Table~\ref{tab:dataset_stats}, sequences 00--07 and 09--10 are used for training and sequence 08 for validation, while sequences 11--21 are used for online test evaluation.
Although SemanticKITTI does not provide dense pixel-level 2D semantic annotations, we generate 2D supervision by projecting LiDAR point-wise labels onto the image plane using known camera–LiDAR calibration parameters. Specifically, each labeled 3D point is mapped to its corresponding pixel coordinate, producing sparse 2D semantic labels. These projected labels are used solely for auxiliary supervision during training and are not required at inference time.

% -------------------------
% ABLATION TABLE
% -------------------------
\begin{table}[!tp]
\centering
\caption{Ablation study on the {SemanticKITTI} validation set. \newline Note: \ding{51}  - included, \ding{55} - omitted}
\label{tab:ablation}
%\scriptsize
\setlength{\tabcolsep}{3pt}
\renewcommand{\arraystretch}{1.1}
\begin{tabular}{lccc}
\toprule
{Configuration} & {MSCA} & {KD} & {mIoU (\%)} \\
\midrule
Baseline (LiDAR only) & \ding{55} & \ding{55} & 67.0 \\
\quad + Contrastive & \ding{51} & \ding{55} & 67.8 \\
\quad + KL-divergence + Contrastive & \ding{51}  & \ding{55} & 68.4 \\ \rowcolor{green!15}
 \texttt{xModel-KD} & \ding{51}  & \ding{51}  & {69.1} \\
\bottomrule
\end{tabular}
\end{table}

% \subsubsection{Evaluation Metrics}
% We report mean Intersection-over-Union (mIoU) as the primary metric and overall accuracy (Acc) when applicable.
% For SemanticKITTI, mIoU is computed over all 19 semantic classes.

\subsection{Implementation Details}

The 3D backbone is SPVCNN and the 2D backbone is ResNet50 pretrained on ImageNet.
We use $S{=}4$ scales and feature dimension $d_h{=}64$.
We train with a total batch size of 16 for 64 epochs. we use cosine learning rate decay and standard point cloud augmentations.
Inference uses only the 3D stream; all 2D and cross-modal components are removed at test time. For performance evaluation, we report mean Intersection-over-Union (mIoU) as the primary metric and overall accuracy (Acc) when applicable.
%For SemanticKITTI, mIoU is computed over all 19 semantic classes.

\subsection{Quantitative Analysis}
\label{sec:quant_results}

\subsubsection{Overall Performance}
\label{sec:ablation}

Table~\ref{tab:semantickitti} presents a comparative evaluation on SemanticKITTI. 
Compared to the LiDAR-only baseline, \texttt{xModel-KD} achieves improved overall performance. 
Among the competing methods, it ranks second behind 2DPASS~\cite{yan2022_2dpass}. 
In terms of model complexity, however, \texttt{xModel-KD} is considerably more efficient, whereas 2DPASS employs a substantially larger architecture with 45.6M parameters, \texttt{xModel-KD} contains only 1.93M parameters over 23× fewer parameters. This highlights a highly favorable performance–complexity trade-off. Such efficiency makes \texttt{xModel-KD} particularly attractive for real-world deployment scenarios, including resource-constrained and industrial environments where memory footprint, computational cost, and inference latency are critical factors

\subsubsection{Component Analysis}

An ablation study is conducted to assess the contributions of (i) multi-scale contrastive alignment, (ii) KL-divergence-based knowledge distillation. As reported in Table~\ref{tab:ablation}, each component provides measurable performance improvements. The complete model achieves the best overall results, outperforming the baseline by $2\%$.

\subsubsection{Sensitivity to Loss Weights}

To evaluate the influence of the loss weights in the multi-task objective, we conduct a sensitivity analysis on $\lambda_{\text{contrast}}$ and $\lambda_{\text{KD}}$, as summarized in Table~\ref{tab:weights}. Recall that $\lambda_{\text{contrast}}$ and $\lambda_{\text{KD}}$ control the relative contribution of the cross-modal contrastive alignment loss $\mathcal{L}_{\text{contrast}}$ and the knowledge distillation loss $\mathcal{L}_{\text{KD}}$ in the total loss, $\mathcal{L}_{\text{total}}$, defined earlier in \eqref{eq-multiloss}. 
The analysis shows that the \texttt{xModel-KD} remains stable across a wide range of weight values, indicating that performance is not overly sensitive to precise tuning. Optimal performance is achieved with 0.1 set for both $\lambda_{\text{contrast}}$ and $\lambda_{\text{KD}}$, demonstrating that both contrastive alignment and knowledge distillation contribute meaningfully.

% -------------------------
% LOSS WEIGHTS TABLE (OPTIONAL)
% -------------------------
\begin{table}[!t]
\centering
\caption{Sensitivity analysis of weight parameters in the multimodal loss used for \texttt{xModel-KD}}
\label{tab:weights}
%\footnotesize
\setlength{\tabcolsep}{4pt}
\renewcommand{\arraystretch}{1.1}
\begin{tabular}{cccc}
\toprule
$\lambda_{\text{contrast}}$ & $\lambda_\text{KD}$ & mIoU (\%) & Acc (\%) \\
\midrule
0.05 & 0.05 & 68.2 & 92.6 \\
0.10 & 0.05 & 68.7 & 92.8 \\
\rowcolor{green!15}
0.10 & 0.10 & {69.0} & {93.0} \\
0.20 & 0.10 & 68.8 & 92.9 \\
\bottomrule
\end{tabular}
\end{table}

% \subsection{Qualitative Results}
% \label{sec:qualitative}

% Table~\ref{tab:semantickitti} shows qualitative comparisons on SemanticKITTI.
% Compared to the LiDAR-only baseline, \texttt{xModel-KD} yields cleaner boundaries and improved recognition of small or thin structures.

% -------------------------
% QUALITATIVE FIGURE PLACEHOLDER
% -------------------------
% \begin{figure}[t]
% \centering
% % Replace these four images with your own exports.
% % (a) RGB, (b) GT, (c) Baseline, (d) Ours
% % \includegraphics[width=0.24\textwidth]{figs/qual_rgb.png}\hfill
% % \includegraphics[width=0.24\textwidth]{figs/qual_gt.png}\hfill
% % \includegraphics[width=0.24\textwidth]{figs/qual_base.png}\hfill
% % \includegraphics[width=0.24\textwidth]{figs/qual_ours.png}
% \caption{\textbf{Qualitative comparison} (mock placeholders). From left to right: RGB image, ground truth, LiDAR-only baseline, and our xModalKD prediction. xModalKD improves boundary consistency and small-object recognition while maintaining LiDAR-only inference.}
% \label{fig:qualitative}
% \end{figure}

\subsection{Qualitative Results}
\label{sec:qualitative}

{Fig.~\ref{fig:qualitative} presents qualitative comparisons between the ground truth and our model predictions in the bird’s-eye view (BEV) on the SemanticKITTI validation set. The results show that our model produces predictions that closely match the ground truth, demonstrating accurate semantic segmentation performance. In particular, the predicted regions exhibit consistent spatial structures. These qualitative observations are consistent with the quantitative results, further confirming the effectiveness of our approach.

% \begin{figure*}[!tp]
% \centering

% % Row 1
% \includegraphics[width=0.32\textwidth]{ground.jpeg}
% \hspace{2pt}
% \includegraphics[width=0.32\textwidth]{prediction.jpeg}

% \vspace{4pt}

% % Row 2
% \includegraphics[width=0.32\textwidth]{ground_79.jpeg}
% \hspace{2pt}
% \includegraphics[width=0.32\textwidth]{ground_79_pre.jpeg}

% \caption{Qualitative results on SemanticKITTI sequence 08.
% Top: sample 000000. Bottom: sample 000079.
% For each pair, left shows ground truth and right shows predictions from \texttt{xModel-KD}.
% Colors: car (blue), motorcyclist (red), road (cyan), parking (orange),
% sidewalk (lime), building (purple), fence (gold), vegetation (sky blue),
% guard-rail (violet), terrain (deep orange), pole (yellow-green), traffic-sign (sea green).}
% \label{fig:qualitative}
% \end{figure*}

\begin{figure*}[!tp]
\centering
\setlength{\tabcolsep}{15pt}
\begin{tabular}{cc}
    Ground Truth & \texttt{xModel-KD}'s Prediction \\
   \includegraphics[width=0.32\textwidth]{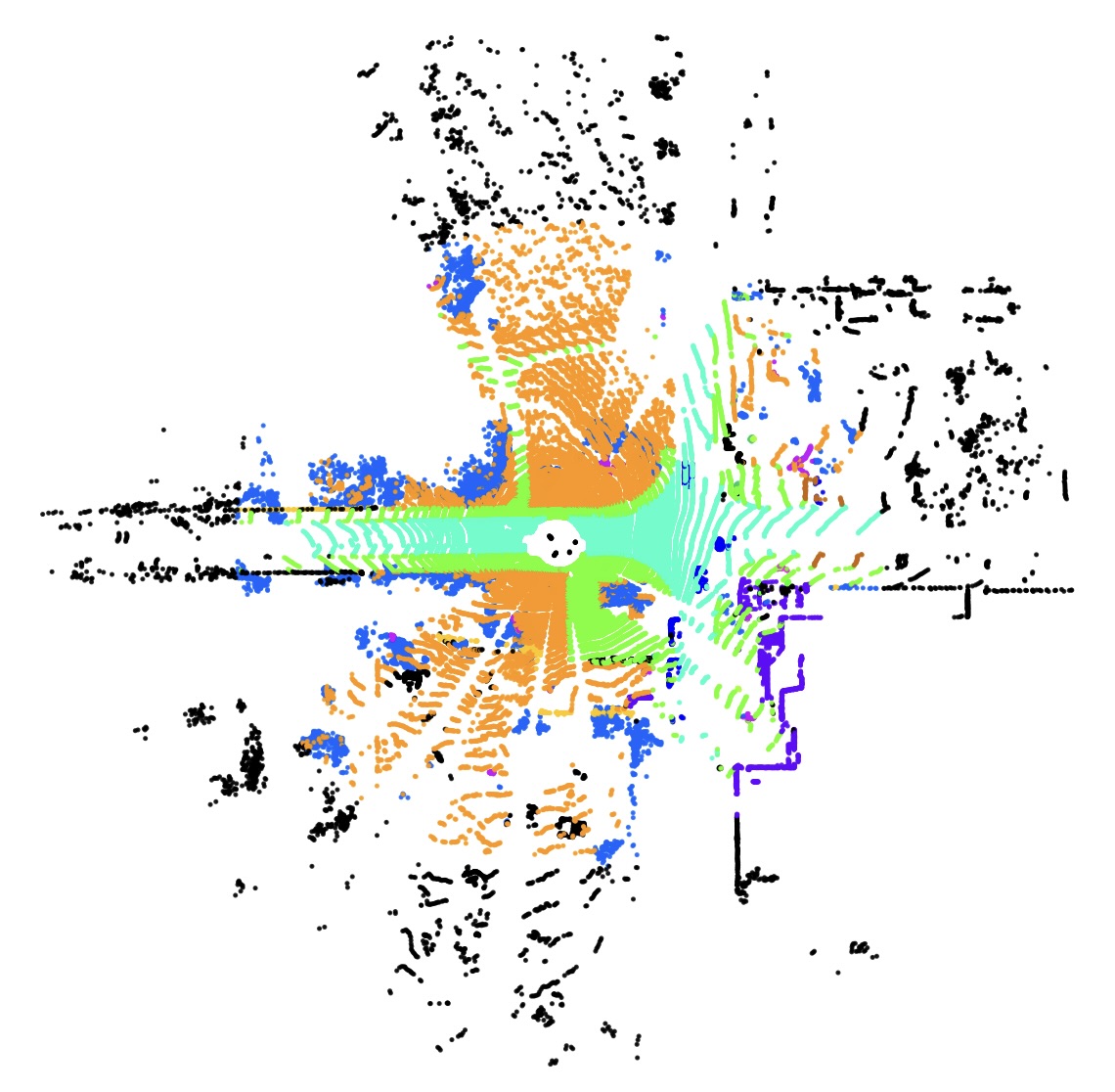} & 
   \includegraphics[width=0.32\textwidth]{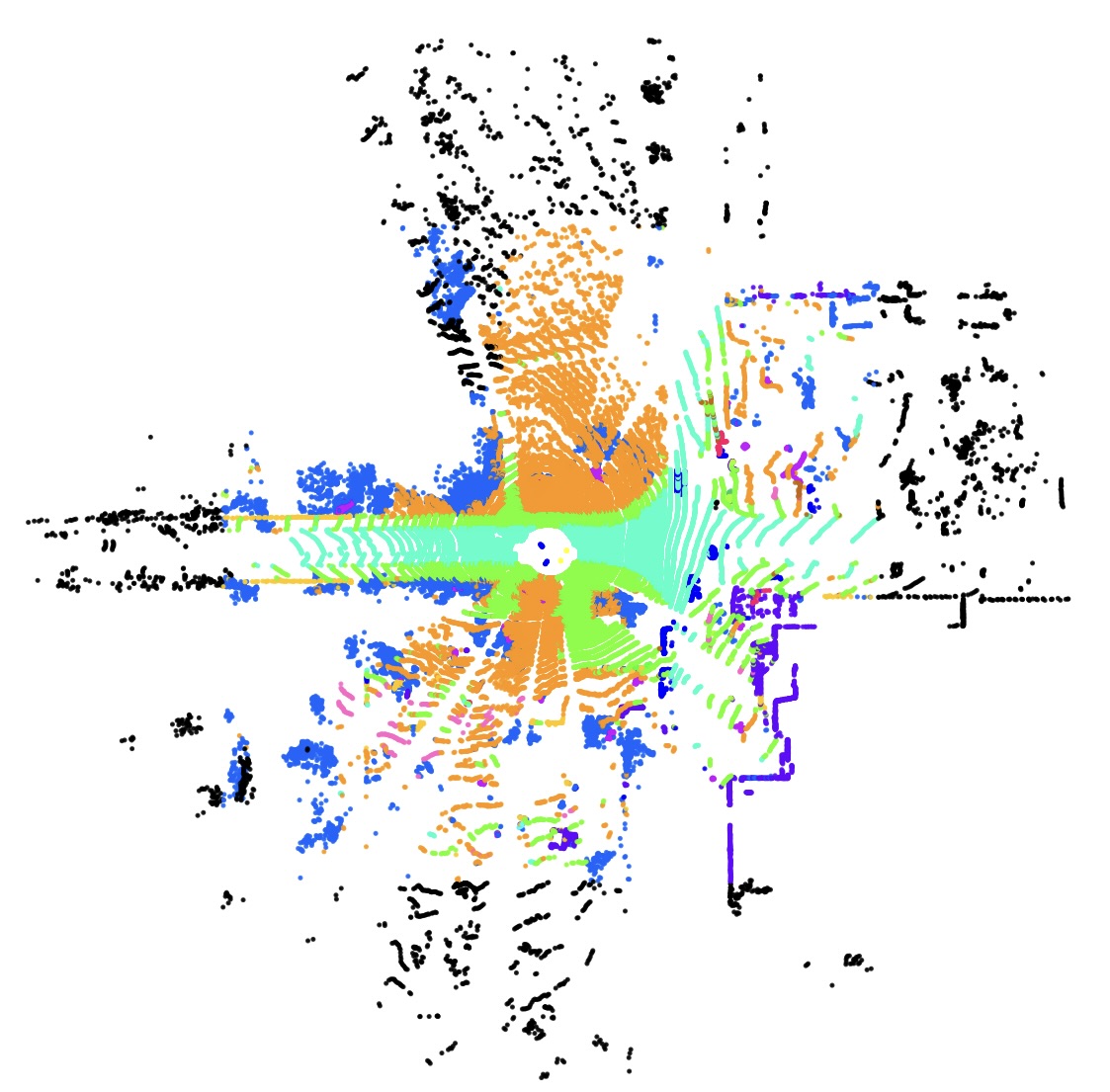} \\
   \multicolumn{2}{c}{\footnotesize For the input sample \texttt{000000.bin} from SemanticKITTI validation set.} \\
   %================ Row 2 =================%
    \includegraphics[width=0.32\textwidth]{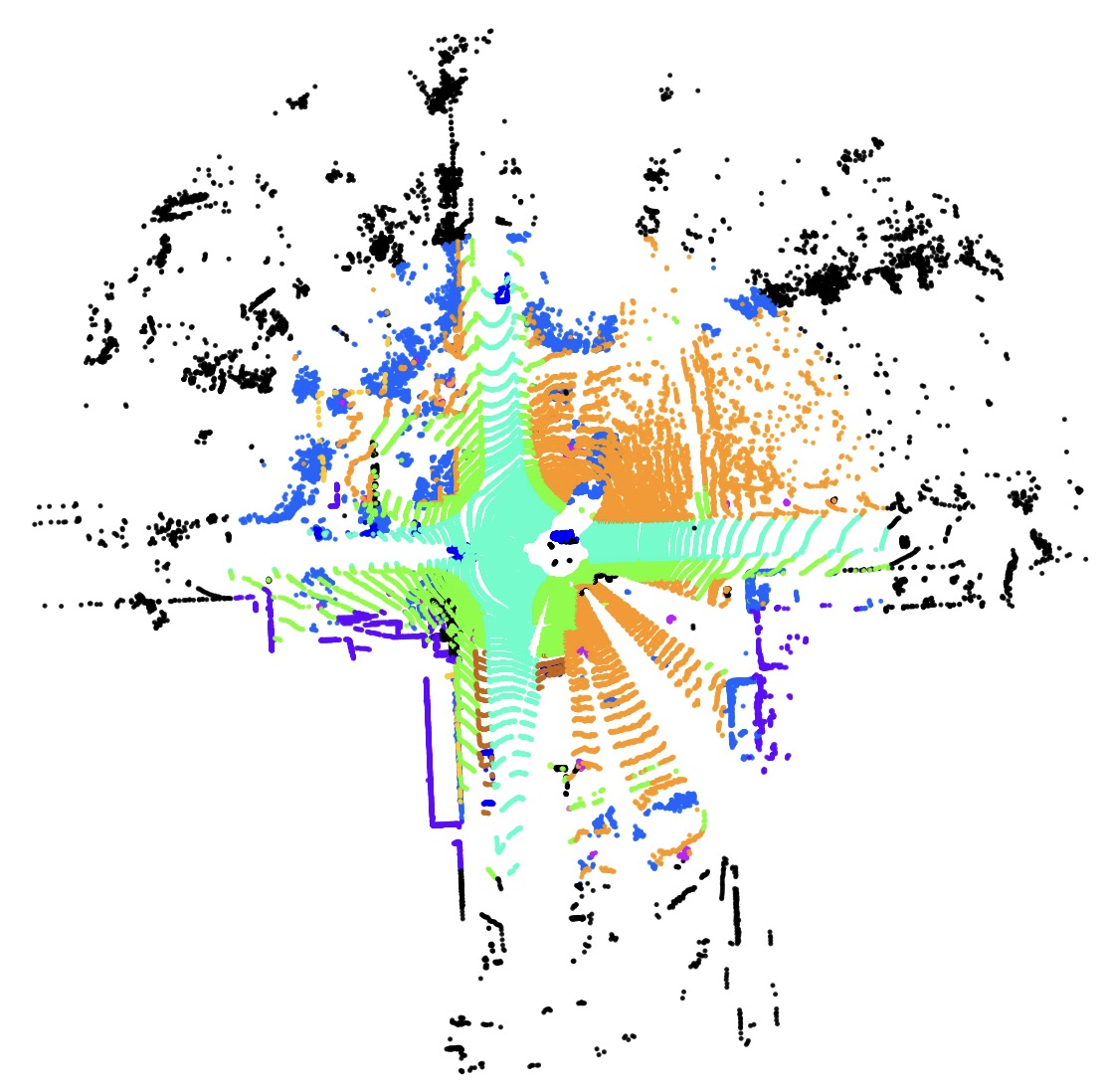} &
    \includegraphics[width=0.32\textwidth]{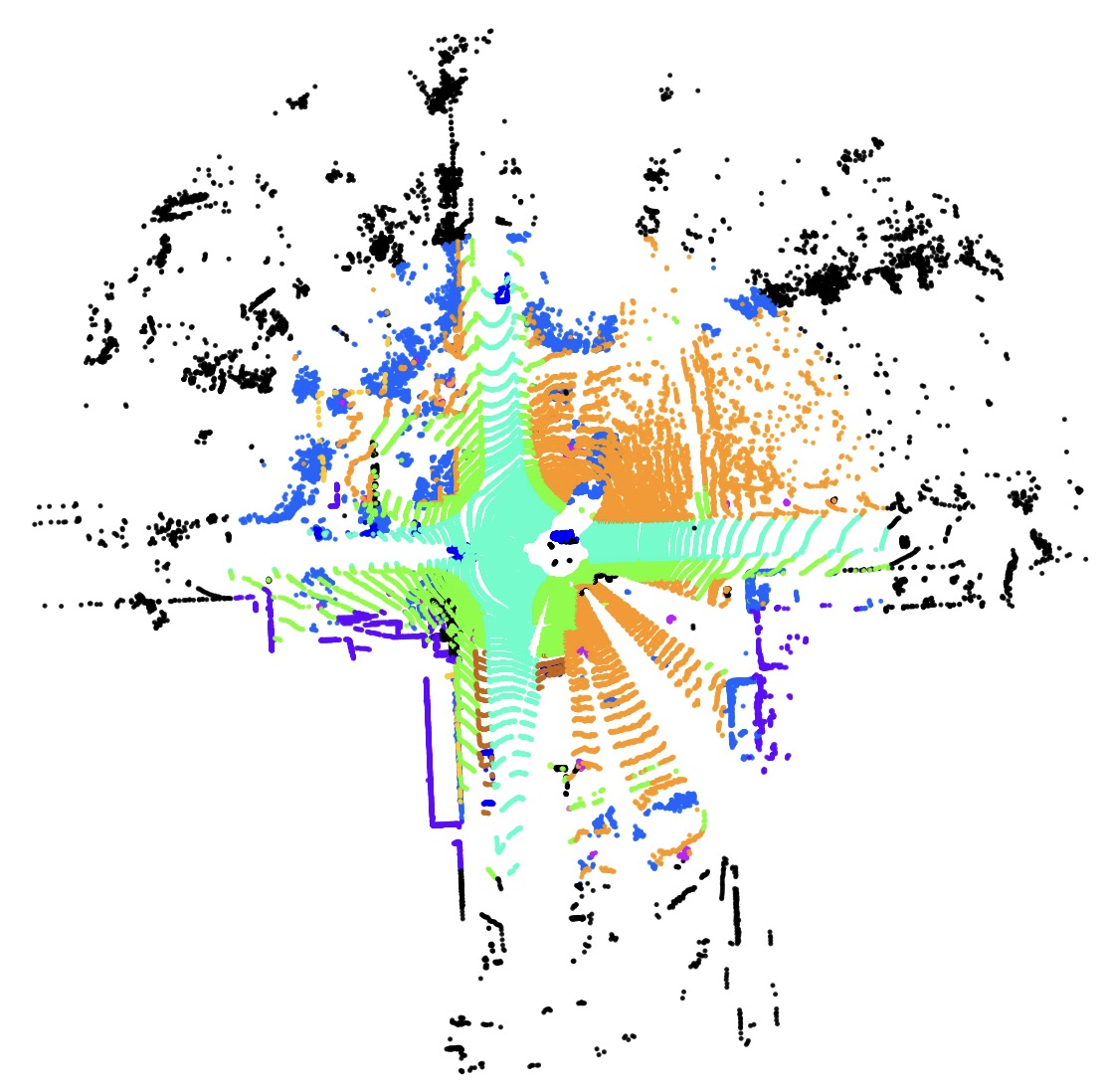} \\
   \multicolumn{2}{c}{\footnotesize For the input sample \texttt{000079.bin} from SemanticKITTI validation set.}\\
\end{tabular}

\vspace{0.2cm}
{\footnotesize\underline{Legend}: %{\color{blue}$\bullet$} car, 
% {\color{red}$\bullet$} motorcyclist,  
{\color[RGB]{152,255,152}$\bullet$} road,
{\color{black}$\bullet$} noise, 
{\color{orange}$\bullet$} parking, 
{\color{lime}$\bullet$} sidewalk,  
{\color{purple}$\bullet$} building, 
{\color{yellow}$\bullet$} fence, 
% {\color{SkyBlue}$\bullet$} vegetation, 
% {\color{Orange}$\bullet$} terrain, 
% {\color{green!60!yellow}$\bullet$} pole, 
{\color{SeaGreen}$\bullet$} traffic-sign. }
\caption{Qualitative results on SemanticKITTI sequence 08. For each pair, left shows ground truth and the right shows predictions from \texttt{xModel-KD}.}
\label{fig:qualitative}

\end{figure*}

\section{Conclusion}

The proposed \texttt{xModel-KD}, a training-only cross-modal knowledge distillation framework for LiDAR semantic segmentation that transfers rich semantic priors from a frozen 2D teacher into a compact 3D backbone while preserving zero inference overhead. Unlike conventional multi-modal fusion methods that require images and heavy fusion modules at test time, \texttt{xModel-KD} decouples cross-modal learning from deployment by leveraging multi-scale contrastive alignment and KL-based distillation during training only. 

Extensive experiments on SemanticKITTI demonstrate that \texttt{xModel-KD} consistently performs across various object classes. %, with ablation studies confirming the complementary contributions of multi-scale contrastive alignment, gated fusion, and prediction-level distillation. Importantly, after training, the entire 2D branch and cross-modal components are removed, resulting in a pure 3D model that maintains enhanced semantic representations without additional runtime cost.
Beyond accuracy improvements, \texttt{xModel-KD} is lightweight in terms of parameters and computational complexity, making it suitable for real-time deployment on resource-constrained platforms. This efficiency, combined with LiDAR-only inference, makes the framework particularly attractive for practical applications such as autonomous systems, warehouse robotics, and indoor industrial inspection. %, where low latency and memory footprint are critical.
Nevertheless, it relies on accurate LiDAR--camera calibration and point--pixel correspondences during training, and its generalization to strongly misaligned or degraded sensor conditions requires further investigation. Overall, \texttt{xModel-KD} demonstrates that training-time cross-modal knowledge transfer is an effective and deployment-efficient strategy for enhancing 3D semantic segmentation without sacrificing inference efficiency.

\bibliographystyle{IEEEtran}
\bibliography{references}

\end{document}